\documentclass[letterpaper, 10 pt, conference]{ieeeconf}  % Comment this line out if you need a4paper

\IEEEoverridecommandlockouts                              % This command is only needed if 
\usepackage{soul}
\usepackage[tracking=true]{microtype}
\usepackage{amsmath} % assumes amsmath package installed
\usepackage{amssymb}  % assumes amsmath package installed
\usepackage{graphicx}
\usepackage{epsfig} % for postscript graphics files
\usepackage{times} % assumes new font selection scheme installed
\usepackage{bm}
\usepackage{xspace}
\usepackage{graphicx}
\usepackage{booktabs}
\usepackage{float}
\usepackage{afterpage}
\usepackage{url}
\usepackage{paralist}
\usepackage{dsfont}
\usepackage{mathrsfs}
\usepackage[linesnumbered,ruled,vlined]{algorithm2e} 
\usepackage{algorithmic} 
\usepackage{comment}

\usepackage[dvipsnames]{xcolor}
\usepackage[font={footnotesize}]{caption}
\usepackage{booktabs} % for professional tables
\usepackage{tikz}
\usetikzlibrary{positioning,shapes.geometric,arrows,fit,shapes.symbols,svg.path,tikzmark,calc}
\usepackage{textcomp}
\usepackage{listings}
\usepackage{subcaption}
\pgfdeclarelayer{background}
\pgfdeclarelayer{foreground}
\pgfsetlayers{background,main,foreground}
\usepackage{siunitx}
\usepackage{adjustbox}
\usepackage{array}
\usepackage{multirow}
\usepackage{booktabs}
\usepackage{tabularx}

\usepackage{siunitx}
\usepackage{tikz}
\usetikzlibrary{positioning,arrows.meta}

\usepackage[table]{xcolor}

\let\labelindent\relax %https://tex.stackexchange.com/questions/170772/command-labelindent-already-defined

\usepackage[inline]{enumitem} % for inline lists, e.g. \begin{enumerate*}[label=\roman*)]  See https://tex.stackexchange.com/questions/146306/how-to-make-horizontal-lists

\usepackage{pifont}
\definecolor{darkpink}{RGB}{194,24,91}

\usepackage[hidelinks]{hyperref}
\usepackage[backend=biber,
            hyperref=true,
            isbn=false,
            doi=false,
            backref=false,
            style=ieee,
            natbib=true,%compatibility aliases
            mincitenames=1,
            maxcitenames=1,
            minbibnames=12,
            maxbibnames=12,
            citestyle=numeric-comp,
            sorting=none,%none
            block=none]{biblatex} 
\renewcommand{\bibfont}{\small}
\SetCommentSty{mycommfont}

\SetKwInput{KwInput}{Input}                % Set the Input
\SetKwInput{KwOutput}{Output}    

\definecolor{codegray}{RGB}{238,238,238}
\definecolor{gapblue}{RGB}{222,232,247}
\definecolor{humangreen}{RGB}{226,239,224}
\definecolor{privpurple}{RGB}{235,226,242}

\title{\bf VABAR: A Variational Automation Benchmark \\for Agentic Robotics}

\author{
\normalfont
    Shuangyu Xie$^{3,*}$,
    Kaiyuan Chen$^{3}$,
 	Uksang Yoo$^{3}$, 
    Xinyue Zhu$^{3}$, 
    Meadow Shen$^{3}$, 
    Jaimyn Drake$^{3}$,\\ 
    Eugen Solowjow$^{1}$, 
    Lionel Gulich$^{2}$, 
    Alexander Millane$^{2}$, 
    Rafael Cathomen$^{2}$, 
    Tuur Stuyck$^{2}$, \\
    Sangeeta Subramanian$^{2}$, 
    Yichao Pan$^{2}$, 
    Bruce McGowan$^{2}$, 
    Michael Benedict$^{2}$, 
    Linxi Fan$^{2}$, 
    Yuke Zhu$^{2}$, \\
    Spencer Huang$^{2}$, 
    Ken Goldberg$^{3}$ \\[0.5em] 
    $^{1}$ Siemens \qquad
    $^{2}$ NVIDIA \qquad
    $^{3}$ University of California, Berkeley \qquad  $^{*}$ equal contribution \\[0.5em]
    \small{ Project Website: \textcolor{darkpink}{\nolinkurl{https://berkeleyautomation.github.io/vabar-website/}}}
}

\begin{document}

\twocolumn[{%
\renewcommand\twocolumn[1][]{#1}%
\maketitle
\vspace{-20pt}
\begin{center}
    \centering
    \captionsetup{type=figure}
    \includegraphics[width=\linewidth]{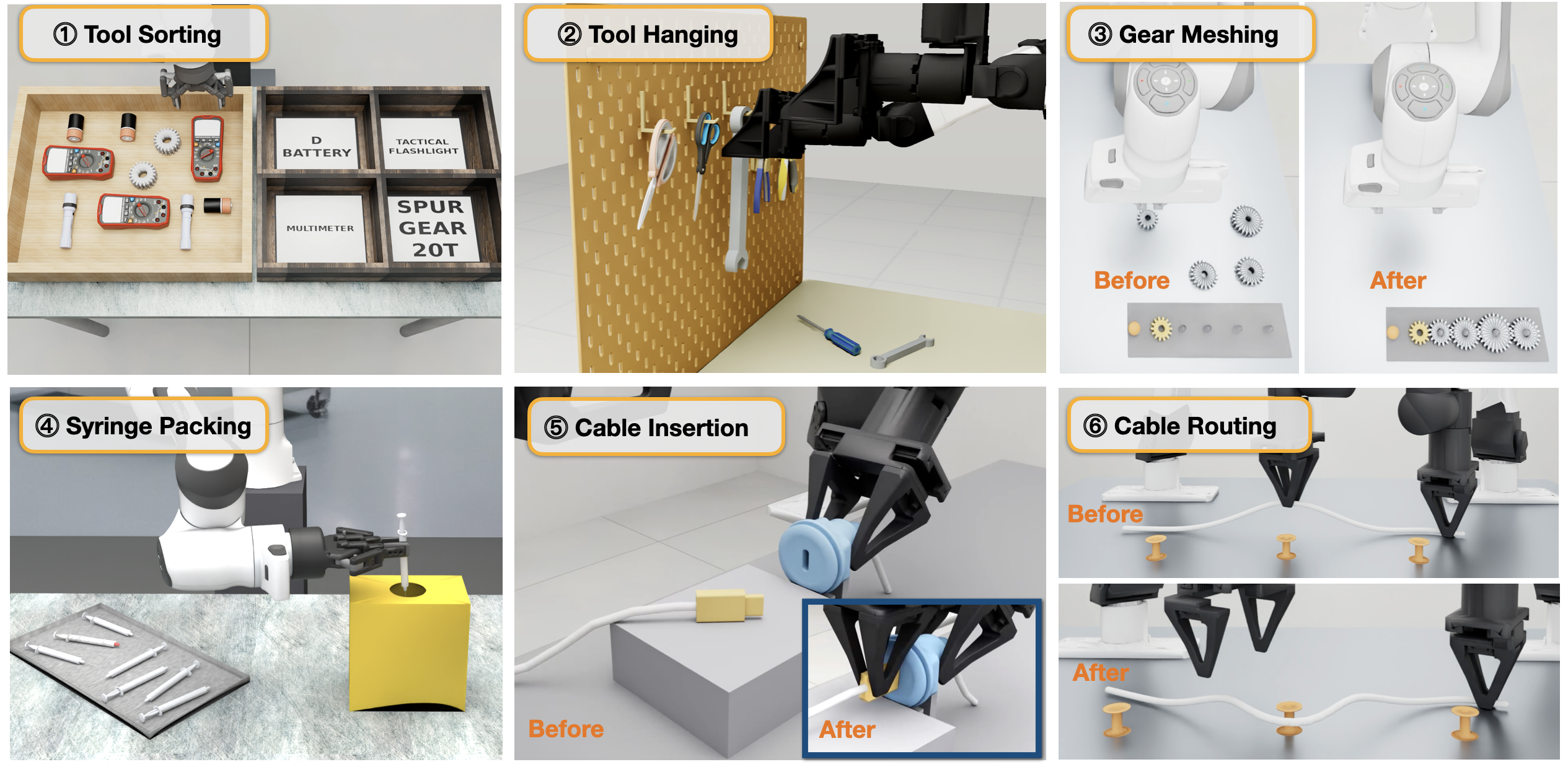}
    \captionof{figure}{Overview. Inspired by industrial applications, VABAR includes six manipulation task classes: (1) Tool Sorting, (2) Tool Hanging, (3) Gear Meshing, (4) Syringe Packing, (5) Cable Insertion, and (6) Cable Routing. The benchmark includes both single-arm and bimanual robot embodiments and evaluates manipulation under specific tiers of geometric and physical variations. VABAR is open access and runs on GPU workstations and on Isaac Lab Arena.}
    \label{fig:scene-title}
\end{center}%
}]

% \maketitle
\thispagestyle{empty}
\pagestyle{empty}

\begin{abstract}

For robots to operate reliably in commercial and industrial applications, they must persistently perform specific tasks with known variations in object shape and pose. 
We call this \emph{Variational Automation} (VA) to differentiate these settings from fixed automation.
Inspired by recent papers on Agentic Robotics, where pretrained LLM and VLM coding agents combine interpretable robot programming with the open-world adaptability of model free policies, 
we propose a general set of agentic robotics benchmark design criteria that includes accessibility, repeatability, diversity, relevance to agentic robotics and practical applications, physical realizability, resistance to benchmark saturation, and computational efficiency. 
We instantiate these principles in six manipulation task classes spanning industrial, medical, and infrastructure applications: tool sorting, tool hanging, gear meshing, syringe packing, cable insertion, and cable routing. 
We introduce the \emph{Variational Automation Benchmark for Agentic Robotics}~({VABAR}), an open, easily replicable, simulation-based manipulation benchmark to systematically evaluating agentic robotics systems.
Built fully on open-source tools, VABAR features detailed implementations using the Newton physics engine, standardized task variations, difficulty tiers, success metrics, and a library of 33 generalizable skills with human-authored programmatic reference solutions.
We provide initial baseline evaluations using an open agentic manipulation harness, Graph-as-Policy (GaP), and Isaac Lab Arena. 
\end{abstract}

\section{Introduction}

For robots to operate reliably in commercial and industrial applications, they must solve
\emph{Variational Automation} (VA) tasks, 
where operational objectives and workspace persist with structured variations in objects, layouts, task parameters, and execution conditions across instances.
In \emph{fixed automation}, robots repeatedly execute carefully engineered programs in highly structured environments, often achieving high reliability and throughput but requiring substantial human effort to design, deploy, and retune when conditions change~\cite{goldberg2026generalists}.
In comparison, generalist robotics aims to perform a broad range of open-ended tasks, but current model-free end-to-end approaches struggle to achieve the speed and reliability required for commercial and industrial adoption.
VA tasks lie between fixed automation and fully generalist robotics.

Recent advances in \emph{agentic robotics} provide a promising approach to implementing VA by combining the reliability and interpretability of classical robot programming with the adaptability of learned policies. 
Rather than directly mapping observations to low-level actions through a monolithic model-free policy, agentic robotic systems use multiple LLM agents to reason about task instances, generate and modify executable robot programs, and invoke perception, planning, and control modules named skills.

Existing robot manipulation benchmarks (e.g., LIBERO~\cite{libero}, ManiSkill~\cite{maniskill2}, and SimplerEnv~\cite{simplerenv}) evaluate the robustness and generalization of language-conditioned deep visuomotor policies by offering simulation environments, human-teleoperated training episodes, and success metrics across new scene configurations. 
Many of these have recently been saturated by systems that achieve very high success rates and they primarily focus on household pick-and-place tasks. 

We introduce \textbf{VABAR}, a \textbf{Variational Automation Benchmark for Agentic Robotics}. VABAR provides six application-oriented manipulation task classes spanning industrial, medical, and infrastructure settings: tool sorting, tool hanging, gear meshing, syringe packing, cable insertion, and cable routing. 
Each task is defined with a set of instances with explicit variation axes, difficulty levels, and success conditions. For each task, we provide a physics and photorealistic simulation environment, a set of generalizable skills to facilitate execution,  and a human-authored reference solution expressed as a computation graph over the same skill library. The reference solution is executed with the same perception, planning, and execution stack available to coding agents.
% , so it serves as a reference point for what expert composition and parameterization of the skill library can achieve.
% , not as an oracle.

We evaluate VABAR using GaP, an open agentic manipulation baseline~\cite{chen2026gap}, and with Isaac Lab Arena to establish baselines and to study how performance scales with task difficulty and structured variation. 

This paper makes 5 contributions: \begin{itemize}
    \item[1] \textbf{Variational Automation task formulation.} We introduce and formalize \emph{Variational Automation}, a manipulation class between fixed automation and unconstrained general-purpose manipulation, in which persistent operational objectives must be achieved across a structured distribution of task instances.
    \item[2] \textbf{Design principles for agentic robotics benchmarks.}  We define practical criteria for evaluating agentic systems, emphasizing accessibility, repeatability, controlled diversity, application relevance, physical realizability, resistance to saturation, and computational efficiency.
    \item[3] \textbf{VABAR task definition} We instantiate these principles in six application-oriented manipulation task classes---(1) tool sorting, (2) tool hanging, (3) gear meshing, (4) syringe packing, (5) cable insertion, and (6) cable routing---with standardized task variations, difficulty levels, success metrics, and support for single-arm Franka and bimanual YAM embodiments.
    \item[4] \textbf{VABAR simulation, Agentic Skills and Human-authored solutions} For each task, we provide a photorealistic simulation environment,  33 generalizable skills that facilitate solving the tasks, and a human-authored code solution.
    \item[5] \textbf{Initial agentic manipulation baselines.} We provide initial baseline evaluations using an open agentic robotic system and characterize performance as task complexity and instance variation increase, establishing reference results and identifying remaining challenges for agentic manipulation.
\end{itemize}

\section{Related Work}

\subsection{Robot Manipulation Benchmarks}
% \ken{NIST, NIST board}

Robot manipulation benchmarking has roots in early efforts to evaluate robotic assembly. In its discussion of benchmark tests for comparing manipulators, the U.S. National Institute of Standards and Technology  defines benchmark for robotics such as the \textit{Robot Characterization Testing} report~\cite{nist_robot_characterization} references Nof and Paul's work on planning robotic assembly~\cite{nof1980assembly}. The Cranfield Assembly Benchmark~\cite{collins1985benchmark} provided a standardized physical assembly test, while YCB~\cite{ycb} later broadened evaluation through shared object sets and protocols. NIST assembly task boards~\cite{kimble2020nist} further standardized tests of insertion, fastening, and gear meshing. Alongside these physical benchmarks, reusable simulation suites emerged to evaluate manipulation algorithms and learned policies.
 Meta-World~\cite{metaworld} and RLBench~\cite{rlbench} provide diverse manipulation task suites, while ManiSkill~\cite{maniskill,maniskill2} emphasizes generalization across manipulation skills and object geometries. CALVIN~\cite{calvin} studies long-horizon language-conditioned manipulation, and LIBERO~\cite{libero} focuses on knowledge transfer and lifelong robot learning. VIMA-Bench~\cite{vima}, ARNOLD~\cite{arnold}, and VLABench~\cite{vlabench} further evaluate multimodal prompting, language grounding, and generalization. Beyond tabletop manipulation, BEHAVIOR~\cite{behavior,behavior1k}, RoboCasa~\cite{robocasa}, and ManiSkill-HAB~\cite{maniskillhab} consider household and mobile manipulation, while FurnitureBench~\cite{furniturebench} and RoboTwin~\cite{robotwin,robotwin2} emphasize contact-rich, assembly, and bimanual tasks.
 RoboLab~\cite{yang2026robolab} provides language-guided scene and variation generation and can perform sensitivity analysis of robot policies, and RoboDojo~\cite{chen2026robodojo} provides a comprehensive simulation and real evaluation to robot Vision-Language-Action policies. 
 These benchmarks enable systematic evaluation across increasingly diverse tasks, environments, and embodiments, but are primarily organized around general-purpose manipulation, household activities, individual skills, or generalization across distinct tasks. VABAR focuses on  application-oriented automation, where a persistent operational objective must be achieved across structured variations of the same task class over time.

\subsection{Robustness and Variation in Manipulation Evaluation}

A growing class of benchmarks explicitly evaluates robustness under controlled variation and distribution shift. LIBERO-Plus~\cite{liberoplus} introduces perturbations across object layouts, camera viewpoints, robot initial states, language instructions, lighting, textures, and sensor noise; LIBERO-PRO~\cite{liberopro} varies manipulated objects, initial states, task instructions, and environments; and LIBERO-X~\cite{liberox} evaluates increasingly challenging spatial, object, and instruction-level shifts. Rather than treating variation as a perturbation around a nominal task, VABAR defines structured variation as an intrinsic part of a \emph{Variational Automation} task: a persistent operational objective is paired with an expected distribution of variation in objects, geometry, layout, task parameters, and execution conditions. Success therefore requires reliable performance across the intended operating range of an automation application, where variation is expected rather than anomalous.

\subsection{Agentic Robotics and Executable Robot Policies}

Recent work has explored robotic systems in which foundation models reason over task instances, invoke external tools, and generate executable robot behaviors. Code-as-Policies~\cite{codeaspolicies} demonstrated that language models can compose perception and control APIs into executable programs, while subsequent agentic approaches extend this paradigm toward task decomposition, planning, tool use, execution feedback, and iterative refinement. CaP-X~\cite{fu2026capx} extends this paradigm by systematically evaluating and improving coding agents for robot manipulation, emphasizing executable policy generation across diverse manipulation tasks.
Recent approaches iteratively improve robot programs through feasibility checking and execution feedback. GaP~\cite{chen2026gap} operates offline; it explicitly represents policies as computation graphs and refines their structure and parameters through parallel simulation rehearsal and generates ROS2-like graph for repeated online execution. ENPIRE~\cite{enpire} extends agentic policy improvement to repeated real-world trials, and ASPIRE~\cite{aspire} accumulates reusable skills through iterative program repair and exploration.

Agentic robotics is particularly relevant to Variational Automation because agentic systems can retain the interpretability and modularity of classical robot programming while adapting behavior across task instances. However, comparing such systems is difficult because they may differ substantially in internal representations, tools, perception interfaces, planning methods, and controllers. VABAR evaluates externally measurable task objectives and structured variation distributions rather than prescribing a particular agent architecture, enabling generated-code, computation-graph, tool-use, learned-policy, and hybrid systems to be compared under common conditions.

\begin{figure*}
    \centering
    \includegraphics[width=\linewidth]{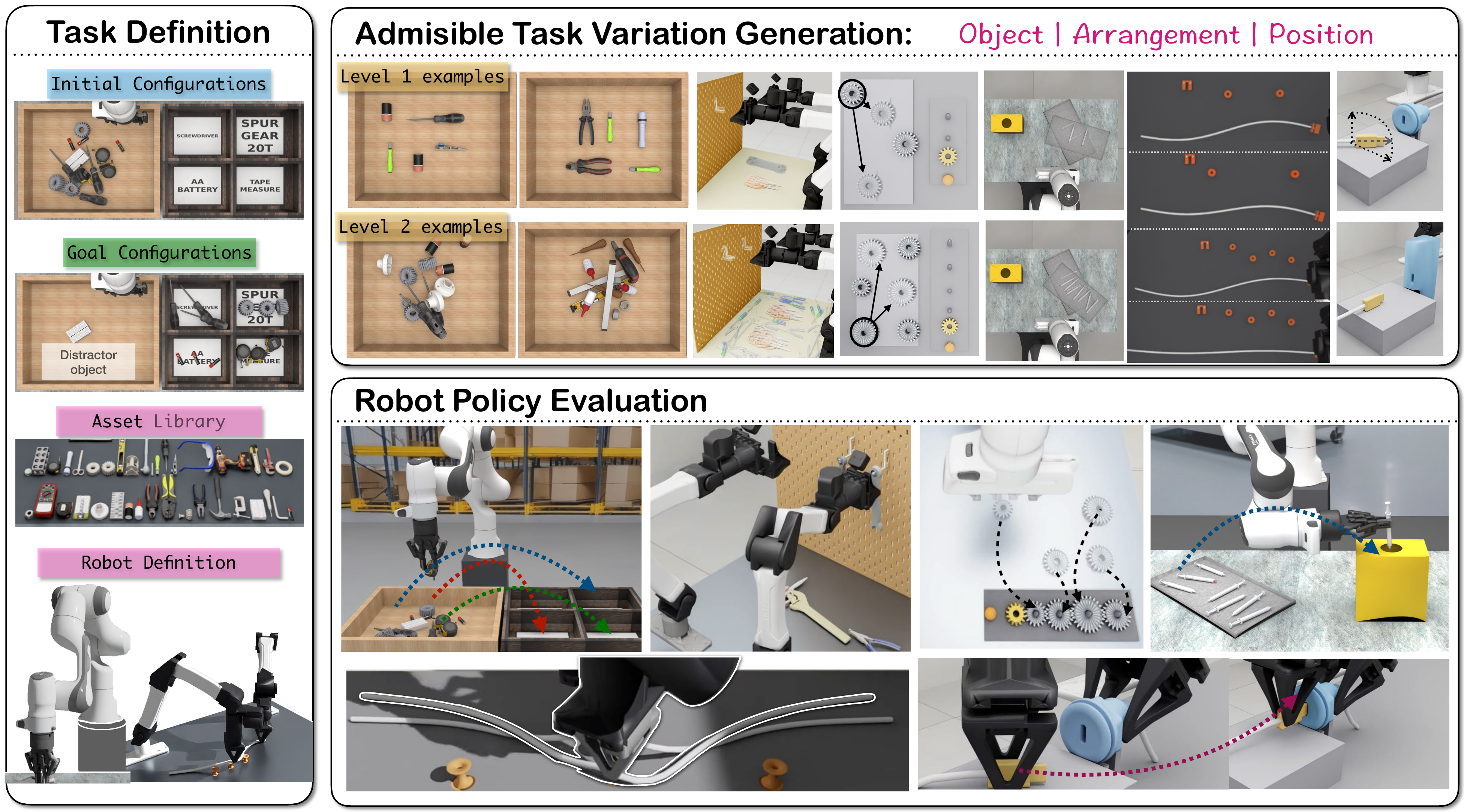}
    \caption{Overview of the VABAR evaluation framework. Left: Tasks are specified by initial and goal configurations, an asset library, and robot definitions. Top right: Admissible task variations span objects, arrangements, and positions, with examples at two levels. Bottom right: Robot policies are evaluated across diverse manipulation tasks, including object sorting, tool hanging, gear assembly, insertion, and cable routing.}
    \label{fig:placeholder}
\end{figure*}

\section{Criteria for Agentic Robotics Benchmarks}
\subsection{Agentic Robotics Benchmark Design Criteria}

% VABAR is designed according to the following principles.
\noindent\textbf{1. Relevance to agentic robotics.}
The benchmark should support evaluating agents that construct,
test, and revise robot programs from task specifications. It
should provide documented robot and tool interfaces for composing
perception, planning, and manipulation capabilities, together with
execution feedback for program and skill refinement during
development. Evaluation should assess reliable task completion
across variations without prescribing a particular program
representation or agent architecture.

\noindent\textbf{2. Application relevance.}
Tasks and variations should reflect realistic operating requirements, with benchmark results predictive of real-world success, cycle time, and intervention needs. 

\noindent\textbf{3. Accessibility and clarity.}
Benchmark environments, task specifications, assets, evaluation
metrics, and variation parameters should be publicly available
and clearly documented so that independent researchers can
reproduce and extend the evaluation.

\noindent\textbf{4. Controlled diversity.}
Each task should specify explicit variation axes and difficulty
levels rather than rely solely on unconstrained randomization.

\noindent\textbf{5. Repeatability.}
Task generation and evaluation should use controlled random
seeds and standardized success criteria to support repeatable
experiments and meaningful comparisons between methods.

\noindent\textbf{6. Computational efficiency.}
Evaluation should be computationally efficient enough to support
repeated experimentation, sensitivity analysis, and large-scale
benchmarking.

\noindent\textbf{7. Durability.}
The benchmark should support systematic extensions of variation
ranges and task complexity to remain informative as robotic
systems improve. Extensions should be versioned to preserve
comparability with previously reported results.

\subsection{Task Definition}

% \tau_i = \operatorname{Instantiate}(\mathcal{T}, O_i)

We define a Variational Automation (VA) task class as
$\mathcal{T}
=
(\mathcal{E},\mathcal{R}, \mathcal{A}, \mathcal{O}, p_{\mathcal{O}}, \mathcal{S}),$
where $\mathcal{E}$ defines environmental coordinate frames,
$\mathcal{R}$ denotes the robot and gripper specification, $\mathcal{A}$ denotes the set of admissible initial assets, each  with physical properties and collision mesh, and $\mathcal{O}$ denotes the space of admissible object configurations. $p_{\mathcal{O}}$ defines a distribution over admissible object configurations, and $\mathcal{S}$ denotes the measurable success condition.
Each task instance configuration $\tau_i =
\left\{
(a_j,\, \mathbf{x}_j)
\right\}_{j=1}^{N_i}
\in \mathcal{O},$
specifies each instantiated  $a_j \in \mathcal{A}$ and its initial pose $\mathbf{x}_j \in SE(3)$ for each of the $N_i$ objects and fixtures in the scene. 

% An individual task instance is generated by sampling
% $$
% \qquad
% \tau_i =
% \operatorname{Instantiate}(\mathcal{T}, O_i), ~\mbox{where}~ O_i \sim p_{\mathcal{O}}. $$
% Thus, a VA task specifies a bounded family of object identities and spatial configurations under a shared robot specification and success condition, rather than a single fixed scene.

Importantly, $\mathcal{O}$ is not defined by unconstrained scene
randomization. Instead, the variation space and its sampling
distribution are designed to represent meaningful changes in the
operating conditions of an automation task.

We use the following
principles to define admissible variations:
{a. Task invariance.}
    All generated instances remain instances of
    the same VA task. {b. Physical feasibility.}
    Generated configurations satisfy physical constraints on object
    support, collision, and geometric consistency, preventing inadmissible initial
    scene configurations.
{c. Robot feasibility.}
    Variations are bounded by the operational workspace of the specified
    robot $\mathcal{R}$. Object and fixture configurations are restricted
    to reachable regions, to
    avoid failures caused solely by physically inaccessible
    configurations.
{d. Controlled difficulty.}
    Each task has levels of difficulty variation regimes.
    Increasing difficulty progressively broadens difficulty for the robot to solve.
{e. Skill preservation.}
    Variations are designed such that instances across the same difficulty levels
    can be solved using the same underlying set of manipulation skills or extensions.

These principles define the admissible configuration space
$\mathcal{O}$ and constrain the support of $p_{\mathcal{O}}$. An
individual task instance is 
$\tau_i \sim p_{\mathcal{O}}.$

\subsection{Task Execution Records}

Task-level success indicates whether an execution achieves its
objective, but does not explain where or why it fails. This is
particularly important for agentic robotics, where execution
feedback may be used to diagnose failures, revise programs, and
introduce recovery behaviors. VABAR therefore records the
execution evidence needed to relate task outcomes to both the
program's decisions and the resulting physical interaction. For agentic robotics, we define the task execution record as decomposition mapping between components.

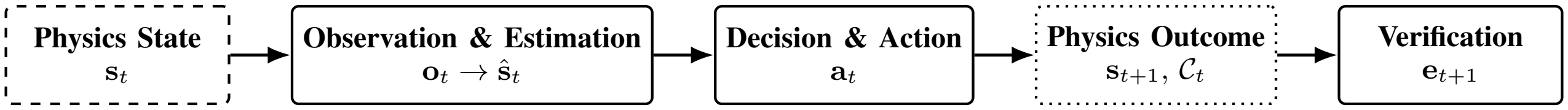
\begin{figure*}[t]
\centering
\begin{tikzpicture}[
    >=Latex,
    font=\small,
    box/.style={
        draw,
        rounded corners=2pt,
        minimum width=2.4cm,
        minimum height=1.05cm,
        align=center,
        line width=0.8pt
    },
    priv/.style={
        box,
        dashed
    },
    priv/.style={
        box,
        dashed
    },
    phys/.style={
    box,
    dotted
    },
    arrow/.style={
        ->,
        line width=0.9pt
    }
]

\node[priv] (state) {
    \textbf{Physics State}\\
    $\mathbf{s}_t$
};

\node[box, right=0.65cm of state] (perception) {
    \textbf{Observation \& Estimation}\\
    $\mathbf{o}_t \rightarrow \hat{\mathbf{s}}_t$
};

\node[box, right=0.65cm of perception] (action) {
    \textbf{Decision \& Action}\\
    $\mathbf{a}_t$
};

\node[phys, right=0.65cm of action] (physical) {
    \textbf{Physics Outcome}\\
    $\mathbf{s}_{t+1},\,\mathcal{C}_t$
};

\node[box, right=0.65cm of physical] (verify) {
    \textbf{Verification}\\
    $\mathbf{e}_{t+1}$
};

\draw[arrow] (state) -- (perception);
\draw[arrow] (perception) -- (action);
\draw[arrow] (action) -- (physical);
\draw[arrow] (physical) -- (verify);

\end{tikzpicture}
\caption{\textbf{Diagnostic execution flow.}
VABAR records observations and estimated state, system actions,
resulting physics states and contacts, and verification outcomes.
Dashed boxes denote privileged simulator state and dotted box denote the physics object contact information used for diagnosis.}
\label{fig:diagnostic_flow}
\end{figure*}

Here, we define $\mathbf{s}_t$ as the true physical state of the
robot and scene, available only to a ``privileged" observer; $\mathbf{o}_t$ is the sensor observation;
$\hat{\mathbf{s}}_t$ is the system's estimated task state;
$\mathbf{a}_t$ is the action produced by the policy, program, or
execution graph; $\mathcal{C}_t$ records physical contacts generated
during execution; and $\mathbf{e}_{t+1}$ denotes the resulting
verification or execution event. We define the following mapping that forms the diagnostic flow, as shown in Fig.~\ref{fig:diagnostic_flow}:
\begin{itemize}
    \item $\mathbf{o}_t \rightarrow \hat{\mathbf{s}}_t$:
    perception or state-estimation failures;

    \item $\hat{\mathbf{s}}_t \rightarrow \mathbf{a}_t$:
    program-generation, planning, skill-selection, or decision failures;

    \item $\mathbf{a}_t \rightarrow
    (\mathbf{s}_{t+1},\mathcal{C}_t)$:
    grasping, motion-execution, collision, contact, or other
    physical-interaction failures;

    \item $(\mathbf{s}_{t+1},\mathcal{C}_t)
    \rightarrow \mathbf{e}_{t+1}$:
    task verification, failure detection, and recovery triggering.
\end{itemize}

The distinction between physical state and contact is important. Object poses and velocities describe the resulting scene
configuration, whereas contact records describe how that
configuration arose through physical interaction. For example, a failed placement may leave an object in an incorrect pose, while its contact history can reveal whether it collided with a fixture, was supported by another object, or became jammed. VABAR records both forms of evidence so that failure attribution can be grounded in the resulting physical execution rather than only in task-level success or software error messages.

% Across instances, the task objective remains unchanged or parameterized within the same application-level objective, while the concrete realization varies:

% \begin{equation}
%     \mathcal{G}_i \approx \mathcal{G}_j,
%     \qquad
%     \theta_i \neq \theta_j.
% \end{equation}

% The variation vector $\theta$ may include factors such as object identity, number of objects, object poses, fixture poses, clutter level, geometric parameters, task sequence, or deformable-object configuration.

% We evaluate task performance using externally measurable metrics. For $N$ evaluation instances, success rate is defined as
% $    \mathrm{SuccessRate}
%     =
%     \frac{1}{N}
%     \sum_{i=1}^{N}
%     \mathbf{1}\left[
%         \mathcal{S}(\tau_i)=1
%     \right].$

% For successful trials, execution efficiency can additionally be characterized using cycle time,

% \begin{equation}
%     T_{\mathrm{cycle}}
%     =
%     \operatorname{median}_{i:\mathcal{S}(\tau_i)=1} T_i.
% \end{equation}

% We report success and execution efficiency separately rather than combining them into a single weighted objective.

% \section{VABAR Benchmark}

\section{VABAR}

VABAR spans six tasks  across rigid-object manipulation, contact-rich assembly, and deformable-object manipulation, covering challenges in semantic perception, geometric reasoning, precision contact, and deformable-object interaction. To support these tasks with a consistent and reproducible benchmark, VABAR integrates four key components: a standardized asset library, representative single- and dual-arm robot embodiments, a unified task specification, and multi-physics simulation support.

\noindent\textbf{Asset Library.}
VABAR provides a  library of over 55 assets, spanning industrial tools, mechanical components, medical supplies, fixtures, connectors, and deformable objects. Assets are normalized in format, scale, appearance, and collision geometry to support controlled object and geometric variation across task instances.

\noindent\textbf{Robot Embodiment.}
VABAR supports the single-arm Franka and bimanual YAM platforms, covering two representative manipulation settings. Franka provides a widely adopted platform for reproducible tabletop manipulation, while YAM enables tasks requiring coordinated bimanual interaction, particularly deformable-object manipulation.

\noindent\textbf{Task Specification.}
VABAR represents tasks declaratively to specify task objects, initial configurations, variations, and success conditions. 

\noindent\textbf{Unified Multi-physics Support.}
VABAR provides unified support for rigid-body and deformable-body simulation and their interactions, with NVIDIA Newton enabling multi-physics dynamics. This allows the benchmark to span rigid manipulation, contact-rich assembly and insertion, and deformable-object tasks such as cable routing.

% \begin{figure}
%     \centering
%     \includegraphics[width=\linewidth]{Image/standardized_asset_library_rtx.png}
%     \caption{VABAR assets. }
%     \label{fig:placeholder}
% \end{figure}

\section{VABAR Tasks}

\begin{figure*}
    \centering
    \includegraphics[width=\linewidth]{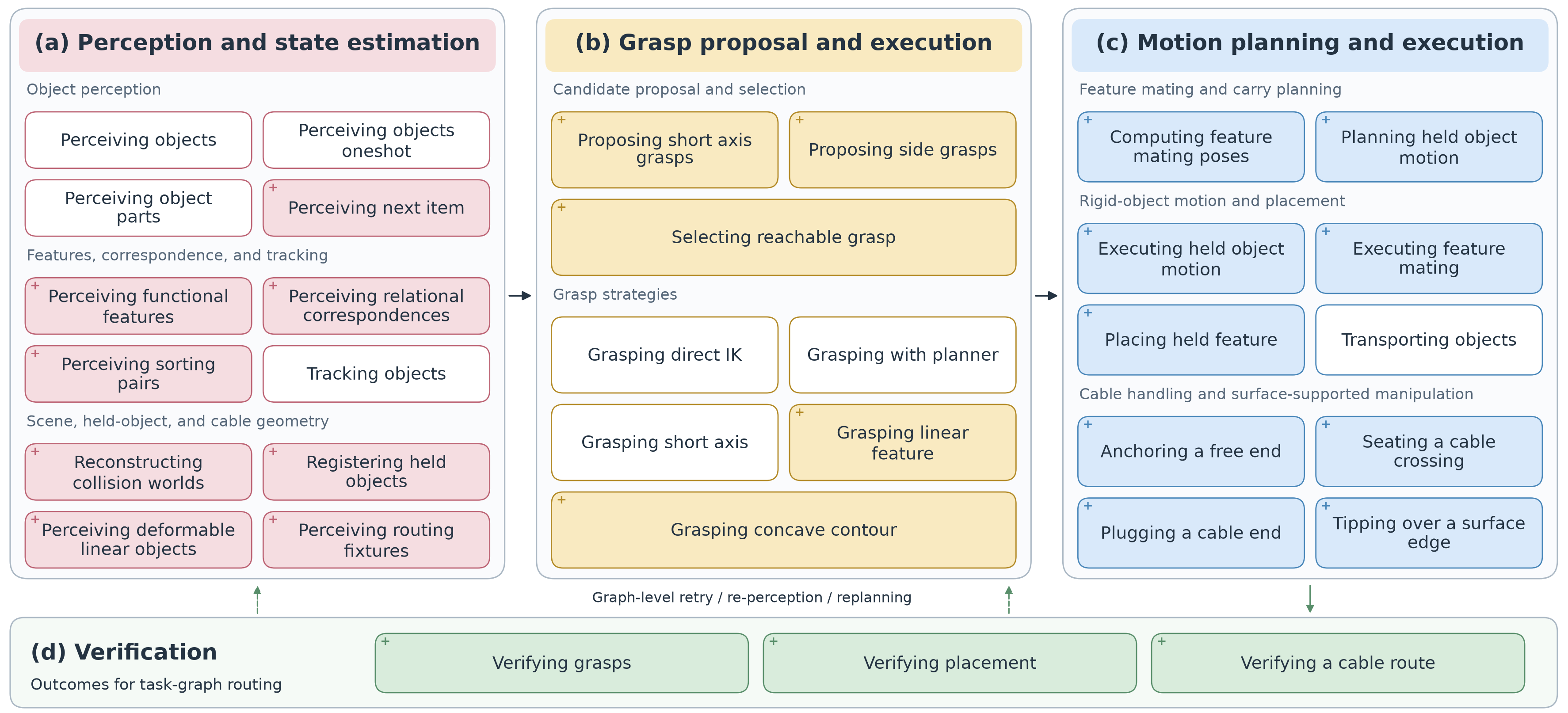}
    \caption{\textbf{VABAR skill library.} The 33 skills span perception,
  grasping, motion planning and execution, and verification. Colored
  cards marked ``+'' identify 25 added skill bundles relative to the
  May Graph-as-Policy library; white cards denote eight retained
  counterparts. Additions support functional-feature manipulation,
  constrained placement, cable handling, and execution checks.}
    \label{fig:vabar-skills}
\end{figure*}

\noindent\textbf{(1) Tool Sorting} The robot must place industrial tools into corresponding bins. Instance variation is introduced by changing the tools, their poses, and tool density. 
% The task contains three difficulty levels:
% \begin{itemize}
%     \item \textbf{Level 1:} singulated tools with no distractors.
%     \item \textbf{Level 2:} overlapping tools
%     % \item \textbf{Hard:} overlapping tools with distractor tools.
% \end{itemize}

% A task instance succeeds when all target objects are placed into the correct bins. Partial progress is measured using the percentage of correctly placed tools.

\noindent\textbf{(2) Tool Hanging}
The robot must pick up one or more industrial tools and hang or insert them into designated hooks, clips, or slots on a pegboard. Tools may include wrenches, screwdrivers, hammers, scissors. Task instance variation is based on tool type, the number of tools, initial tool pose, and holder pose. 
% Difficulty is defined as:
% \begin{itemize}
%     \item \textbf{Level 1:} One tool with a known fixture pose.
%     \item \textbf{Level 2:} multiple tools, multiple fixture types with randomized poses.
%     % \item \textbf{Hard:} multiple tools and fixture types in a cluttered scene with randomized fixture poses.
% \end{itemize}

% Success requires all target tools to be correctly placed into their designated holders and remain stably attached after release. Partial progress is measured using the percentage of correctly placed tools.

% Add to your preamble:
% Preamble:

% Preamble: \usepackage{booktabs,tabularx,array}

\begin{table}[t]
\centering
\caption{Task difficulty levels. 1. TS: Tool Sorting;
2. TH: Tool Hanging; 3. GM: Gear Meshing;
4. SP: Syringe Packing; 5. CI: Cable Insertion;
6. CR: Cable Routing.}
\label{tab:task-difficulty}
\footnotesize
\setlength{\tabcolsep}{3pt}
\renewcommand{\arraystretch}{1.05}
\begin{tabularx}{\columnwidth}{
    @{} l
    >{\raggedright\arraybackslash}X
    >{\raggedright\arraybackslash}X @{}
}
\toprule
\textbf{Task} & \textbf{Level 1} & \textbf{Level 2} \\
\midrule
1 TS & Singulated tools; no distractors
   & Cluttered tools \\
\addlinespace[2pt]
2 TH & One tool; known fixture pose
   & Multiple tools and fixture types; randomized poses \\
\addlinespace[2pt]
3 GM & Assemble two gears
   & Assemble three gears \\
\addlinespace[2pt]
4 SP & Singulated targets
   & Cluttered scenes \\
\addlinespace[2pt]
5 CI & Fixed, visible port; nearby connector with small misalignment
   & Randomized port pose; moderate connector misalignment \\
\addlinespace[2pt]
6 CR & 2--3 widely spaced pegs; short route; low turning angle
   & 4--6 moderately spaced pegs; multiple turns; moderate initial cable variation \\
\bottomrule
\end{tabularx}
\vspace{-10pt}
\end{table}
\noindent\textbf{(3) Gear Meshing}
The task input is gears and axles.
The single-arm robot must pick up multiple gears, assemble them onto a axles, such that they properly mesh, and press a button to activate the mechanism.
The task varies the number of gears, gear size, and initial gear poses. To ensure physically valid instances, scenes can be initialized from a valid assembled configuration before selected gears are removed and repositioned. 
% Difficulty is defined as:
% \begin{itemize}
%     \item \textbf{Level 1:} assemble two gears.
%     \item \textbf{Level 2:} assemble three gears.
%     % \item \textbf{Hard:} assemble four gears, potentially with partially assembled initial configurations and angular variation.
% \end{itemize}

% A trial succeeds when the assembled gear train rotates properly after activation.

\noindent\textbf{(4) Syringe Packing}
The robot must place syringes into a disposal container while avoiding distractor objects. Instance variation includes the number object and disposal box type. 
% Difficulty is defined as:
% \begin{itemize}
%     \item \textbf{Level 1:} singulated target objects with no distractors.
%     \item \textbf{Level 2:} cluttered scenes with diverse syringe types.
%     % \item \textbf{Hard:} cluttered scenes requiring syringes to be pushed through a deformable door flap on the disposal container
% \end{itemize}

% Success requires all syringes to be placed into the disposal box. Partial progress is measured by the percentage of successfully disposed objects.

\noindent\textbf{(5) Cable Insertion}
The YAM robot must pick up a connector, align it with its a USB-C port, and fully insert it. Instance variation includes the position and orientation of the connector and the port. 
% Difficulty is defined as:
% \begin{itemize}
%     \item \textbf{Level 1:} the connector begins near a fixed, clearly visible port with small positional and angular misalignment.
%     \item \textbf{Level 2:} randomized port pose with moderate connector misalignment.
%     % \item \textbf{Hard:} Multiple cables and ports
% \end{itemize}

% Success requires all connector to be fully inserted into the corresponding ports. Partial progress is measured by the percentage of successfully inserted cables.

\noindent\textbf{(6) Cable Routing}
The bimanual YAM robot must route a flexible cable around a specified ordered sequence of cylindrical fixtures mounted on a planar routing board.
The task varies the number, position, and spacing of pegs; the ordered target sequence; cable length; initial cable configuration; cable slack and curvature; endpoint position; and the 3D pose of the routing board.

\definecolor{capgray}{RGB}{238,238,238}
\definecolor{gapblue}{RGB}{222,232,247}
\definecolor{expertgreen}{RGB}{226,239,224}
\definecolor{oraclepurple}{RGB}{235,226,242}

% Preamble: \usepackage{booktabs,multirow,tabularx}
% Keep your existing xcolor/colortbl setup and color definitions.
\begin{table*}[t]
\centering
\caption{Task performance comparison across VABAR tasks and difficulty levels.}
\label{tab:main_results}
\small
\setlength{\tabcolsep}{4pt}
\renewcommand{\arraystretch}{1.12}

\begin{tabularx}{\textwidth}{@{}ll*{8}{>{\raggedleft\arraybackslash}X}@{}}
\toprule
& &
\multicolumn{2}{c}{\cellcolor{expertgreen}\shortstack{\textbf{a-i. GaP}\\[-1pt]\footnotesize(Human w/ Newton)}} &
\multicolumn{2}{c}{\cellcolor{oraclepurple}\shortstack{\textbf{a-ii. GaP}\\[-1pt]\footnotesize(Human w/ ILA)}} &
\multicolumn{2}{c}{\cellcolor{capgray}\shortstack{\textbf{b. Claude}\\[-1pt]\footnotesize w/ Newton}} &
\multicolumn{2}{c}{\cellcolor{gapblue}\shortstack{\textbf{c. GaP}\\[-1pt]\footnotesize w/ Newton}} \\
\cmidrule(lr){3-4}
\cmidrule(lr){5-6}
\cmidrule(lr){7-8}
\cmidrule(lr){9-10}
\textbf{Task} & \textbf{Diff.} &
\textbf{Succ. (\%)} & \textbf{Cyc. (s)} &
\textbf{Succ. (\%)} & \textbf{Cyc. (s)} &
\textbf{Succ. (\%)} & \textbf{Cyc. (s)} &
\textbf{Succ. (\%)} & \textbf{Cyc. (s)} \\
\midrule

\multirow{2}{*}{Tool Sorting}
& Level 1 & \textbf{47.1} & 421.6 & 20.0 & 555.0 & 0.0 & N/A & 26.7 & \textbf{134.2} \\
& Level 2 & \textbf{33.7} & 913.8 & -- & -- & 0.0 & N/A & 13.3 & \textbf{232.3} \\
\addlinespace[3pt]

\multirow{2}{*}{Tool Hanging}
& Level 1 & \textbf{86.7} & 86.1 & 67.5 & 98.3 & 16.7 & 96.2 & 33.3 & \textbf{81.5} \\
& Level 2 & \textbf{81.7} & 158.4 & 50.0 & \textbf{138.3} & 10.0 & 190.8 & 10.0 & 270.9 \\
\addlinespace[3pt]

\multirow{2}{*}{Gear Meshing}
& Level 1 & \textbf{100.0} & 165.9 & 87.5 & 198.9 & 0.0 & N/A & 70.0 & \textbf{156.6} \\
& Level 2 & \textbf{90.0} & \textbf{235.5} & 80.0 & 381.3 & 0.0 & N/A & 60.0 & 284.1 \\
\addlinespace[3pt]

\multirow{2}{*}{Syringe Packing}
& Level 1 & \textbf{87.0} & \textbf{228.8} & 52.5 & 939.0 & 0.0 & N/A & 33.3 & 371.1 \\
& Level 2 & \textbf{63.0} & \textbf{472.3} & 40.0 & 1810.0 & 0.0 & N/A & 12.1 & 652.2 \\
\addlinespace[3pt]

\multirow{2}{*}{Cable Insertion}
& Level 1 & 86.7 & \textbf{78.8} & \textbf{97.5} & 104.9 & 0.0 & N/A & 36.7 & 79.0 \\
& Level 2 & \textbf{30.0} & \textbf{74.2} & 17.5 & 118.8 & 0.0 & N/A & 23.3 & 80.5 \\
\addlinespace[3pt]

\multirow{2}{*}{Cable Routing}
& Level 1 & 46.7 & 116.0 & \textbf{50.0} & 145.9 & 3.3 & 120.7 & 36.7 & \textbf{50.8} \\
& Level 2 & \textbf{46.7} & 189.8 & 5.0 & 151.5 & 0.0 & N/A & 13.3 & \textbf{120.3} \\
\bottomrule
\end{tabularx}

\vspace{3pt}
\begin{minipage}{\textwidth}
\footnotesize
\textit{Metrics:} \textbf{Succ.}: success rate (\%);
\textbf{Cyc.}: median cycle time (s) among successful trials.
\end{minipage}
\end{table*}

\section{VABAR Skills}
  To support the evaluation of agentic robotics systems in the
  VABAR task sandbox, we extend the Modular Open Robot Skill Library(MORSL)~\cite{chen2026gap}
  with reusable procedures for functional-feature manipulation,
  constrained placement, and deformable cable handling
  (Fig.~\ref{fig:vabar-skills}). Building on existing perception,
  grasp generation, and collision-aware planning primitives, the
  extensions provide metric feature localization, source--destination
  correspondence, in-hand registration, and staged feature mating.
  Cable-related skills estimate and track ordered 3D centerlines
  and support anchoring, routing, and terminal placement.

\section{Baseline Experiments}
% VABAR is designed to investigate how agentic robotic systems behave under
% Variational Automation (VA).
We focus on three questions:
(1) how does agentic robotics success rate and cycle time change as the magnitude and dimensionality of
task variation increase?
(2) what are failure modes: perception, grasping, and motion?
(3) can agentic systems reuse skills as
task difficulty changes?

\subsection{Robot Simulation Implementation}
We provide two implementations of VABAR: on Newton and on Isaac Lab-Arena (ILA). The implementations have different texture, lighting, asset and robot implementations.

\noindent \textbf{Newton:} 
We develop VABAR with Newton Physics engine. To determine the success conditions, we extend Behavior Domain Definition Language (BDDL)~\cite{behavior} to represent both rigid and deformable objects as a common specification across the full task suite. 

\noindent \textbf{Isaac Lab-Arena (ILA):}
We provide an implementation of VABAR on top of NVIDIA Isaac Lab-Arena, a framework for composing, executing and analyzing policy evaluation benchmarks with GPU-acceleration. Isaac Lab-Arena provides tools for environment composition, including object placement based on semantic spatial relationships, multiple physics backends including Newton. It also enables running parallel evaluation on GPUs and analysis of policy robustness across environment and task variations. We execute VABAR on a cluster of NVIDIA L40 GPU nodes.

\subsection{Evaluation Protocol}

% A public evaluation manifest
% fixes the instance seeds in advance, ensuring that all conditions encounter
% identical task configurations.

We report the success rate, cycle time, and failure mode rates. Success is determined by the dependency structure of each task. For
tasks composed of repeated manipulation attempts that are independent by
construction, such as sorting or hanging independently sampled objects, each
object-level attempt is treated as a trial and success rate is the fraction of
objects successfully manipulated. This avoids assigning zero success to an
otherwise largely successful multi-object execution.

In tasks such as Gear Meshing, later insertions
are conditioned on the states produced by earlier insertions: inserting one
gear changes the geometry and contact configuration encountered by subsequent
gears. Similarly, multiple cable insertions are treated as separate trials
only when each insertion is independently initialized or reset; otherwise,
their outcomes remain coupled through the evolving task state. We
retain the complete task instance as the evaluation unit for sequential,
state-dependent tasks, while optionally reporting subgoal-level success
(e.g., success for each gear position) as a diagnostic measure. \emph{Cycle time} is the simulated time elapsed from policy start until
success, failure, or timeout.

For each unsuccessful execution, we record the observed failure outcome and assign a primary failure source when supported by execution evidence. For cascading failures, we attribute the execution to the earliest supported failure that explains the subsequent outcome. Failure-source percentages are conditioned on unsuccessful executions.

\subsection{Baselines}
We integrate VABAR with GaP baselines. GaP is a multi-agent harness that
represents a robot policy as a computation graph whose nodes are skills from MORSL~\cite{chen2026gap}.
We evaluate the following systems, labeled as in Table~\ref{tab:main_results}:
(a)~\emph{GaP (Human)}, a manually specified computation graph connecting
MORSL skills, executed on both the Newton and ILA implementations
(Human--Newton and Human--ILA);
(b)~\emph{Claude Code}, a general-purpose coding agent (Claude Code with
Claude Opus~5~\cite{anthropic2026claudeopus5}) without a robotics-specific
harness or skill library;
(c)~\emph{GaP}, where the computation graph is autonomously generated by the
same model and connected to MORSL.
Systems (a) and (c) share the same perception, skill, planning, and execution
modules, so their gap reflects graph composition and parameterization alone.
Each task--difficulty combination contains \(N=30\) independently generated instances per evaluation condition.

% describe GaP, 

\begin{table}[tb]
\centering
\caption{Failure modes across VABAR difficulty levels.}
\label{tab:failure_results}

\small
\setlength{\tabcolsep}{3pt}
\renewcommand{\arraystretch}{1.10}

\begin{tabular}{@{}llrrr@{}}
\toprule
\textbf{Task} &
\textbf{Difficulty} &
\textbf{Perception} &
\textbf{Grasp} &
\textbf{Motion} \\
& & (\%) & (\%) & (\%) \\
\midrule

\multirow{2}{*}{Tool Sorting}
& Level 1   & 37.8 & 51.4 & 10.8 \\
& Level 2 & 56.4 & 28.7 & 14.9 \\
\addlinespace[2pt]

\multirow{2}{*}{Tool Hanging}
& Level 1   & 0.0  & 0.0  & 100.0 \\
& Level 2 & 21.4 & 28.6 & 50.0 \\
\addlinespace[2pt]

\multirow{2}{*}{Gear Meshing}
& Level 1   & N/A & N/A  & N/A \\
& Level 2 & 7.1 & 21.4 & 71.4 \\
\addlinespace[2pt]

\multirow{2}{*}{Syringe Packing}
& Level 1   & 0.0 & 0.0 & 100.0 \\
& Level 2 & 57.9& 0.0 & 42.1 \\
\addlinespace[2pt]

\multirow{2}{*}{Cable Insertion}
& Level 1   & 50.0 & 0.0 & 50.0 \\
& Level 2 & 0.0  & 4.0 & 96.0 \\
\addlinespace[2pt]

\multirow{2}{*}{Cable Routing}
& Level 1   & 27.3 & 45.5 & 27.3 \\
& Level 2 & 5.3  & 31.6 & 63.2 \\

\bottomrule
\end{tabular}

\vspace{2pt}
\begin{minipage}{0.95\linewidth}
\footnotesize
\textit{Failure modes:}
\textbf{Perc.}: perception;
\textbf{Grasp}: grasp generation or execution;
\textbf{Motion}: motion planning or trajectory execution,
including insertion, placement, and routing.
Percentages are conditioned on failures.
% Gear Meshing attribution is per failed gear;
% Syringe Disposal and cable-task attribution is per failed attempt.

\end{minipage}
\vspace{-10pt}
\end{table}

\subsection{Performance}

Table~\ref{tab:main_results} summarizes  performance. Across all evaluated tasks, system performance exhibits a consistent inverse relationship with task difficulty: transitioning from low-variation regimes to multi-object, cluttered environments induces notable performance degradation alongside increased execution cycle times. This drop is particular in precision alignment and deformable object tasks, such as USB-C insertion and cable routing, where introducing geometric variation and tight physical constraints significantly decreases overall task completion. In contrast, discrete item-level manipulation tasks like tool hanging maintain relatively stable per-object success rates across difficulty tiers, though total episode times scale with object count. Overall, execution duration closely reflects task geometry and physical complexity, with single-object hanging tasks completing rapidly while multi-stage routing and cluttered tray operations require substantially longer execution windows. In addition, the same graph running in two different framework (Newton and ILA) has different object texture, scene configuration and lighting.

\section{Discussion}

\noindent\textbf{(a) Agentic harness and skill library enabling better performance.}
Using the same underlying language model, GaP achieves higher success rates than Claude Code in 11 of 12 task--difficulty conditions, supporting the value of a robotics-specific harness coupled with reusable manipulation skills. For example, GaP achieves 70.0\% and 60.0\% success on Level 1 and Level 2 Gear Meshing, respectively, whereas Claude Code achieves no successful trials. However, GaP remains below the human-authored graphs on Newton across all conditions, despite sharing the same perception, skill, planning, and execution stack. This gap suggests that reliable skill composition and parameterization remain important challenges for autonomous policy generation. These comparisons evaluate the combined system; they do not isolate the contributions of the harness and skill library.

\noindent\textbf{(b)~Most errors are due to motion planning.} We decompose unsuccessful GaP executions into perception, grasp, and motion-planning failures. Table~\ref{tab:failure_results} reports failure sources across tasks and difficulty levels. In low-variation settings, perception and grasp generation operate with high reliability across most tasks, leaving motion planning and trajectory execution as the primary source of failure. However, as task variation increases, the dominant failure modes branch based on physical task characteristics. In cluttered scenes and deformable-object manipulation—such as syringe packing and cable routing—perception and grasping degrade significantly due to visual occlusions and unmodeled cable mechanics. Conversely, in precision contact tasks like USB-C insertion, state estimation remains robust under variation, but motion planning and fine insertion alignment overwhelmingly account for nearly all observed failures.

\begin{figure}
    \centering
    \includegraphics[width=\linewidth]{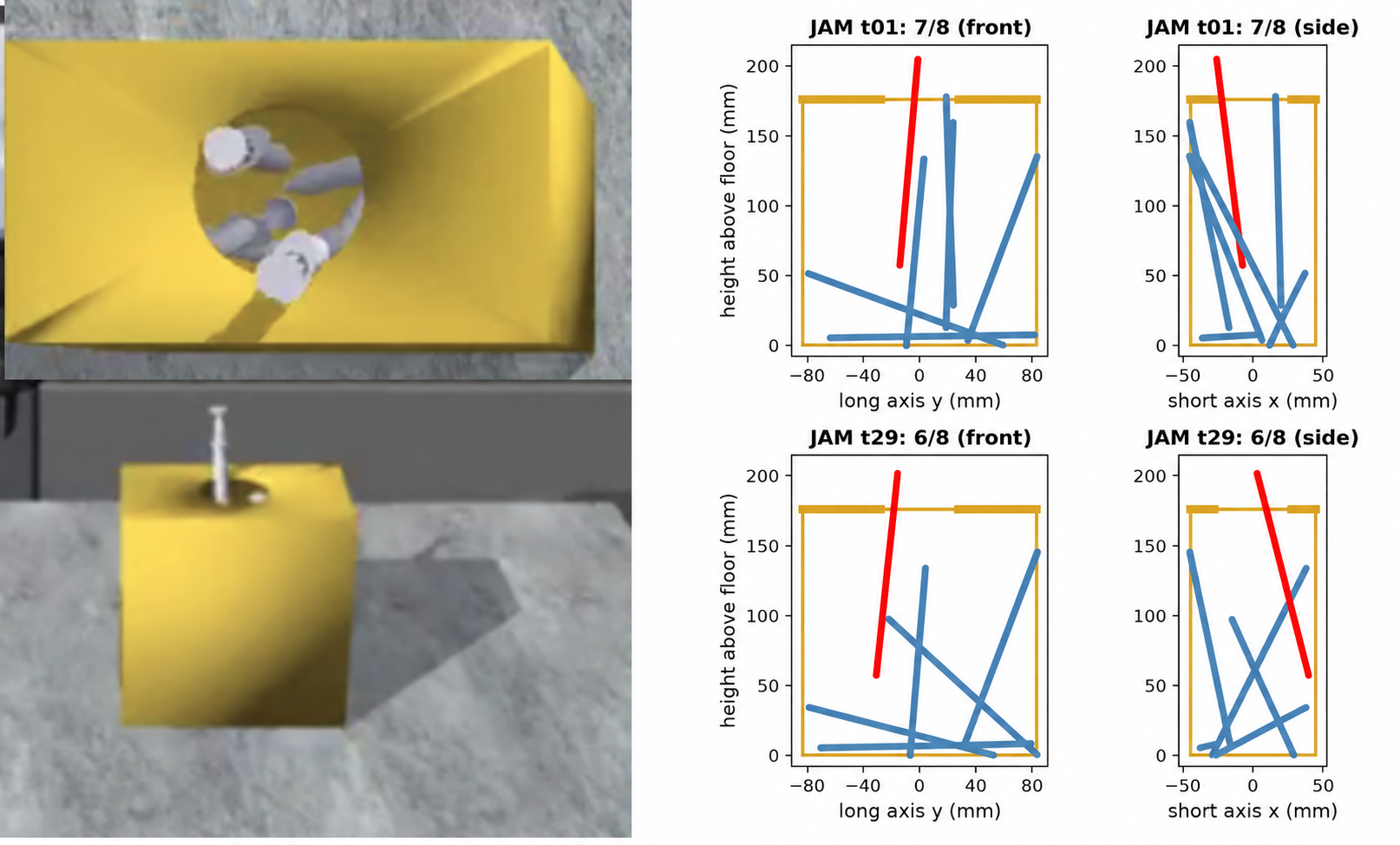}
    \caption{\textbf{Example: Agent inspects Newton physics model to diagnose and fix failure due to syringes jamming inside disposal box .}
Left: simulated syringe-disposal scene with a top-down inset.
Right: Netwon ground-truth front and side internal views of jammed episodes \texttt{t01}
and \texttt{t29}. Red segments highlight syringes straddling the lid plane; blue segments show other syringes.}
    \label{fig:syringe_failure_diagnosis}
    \vspace{-2em}
\end{figure}

\noindent\textbf{(c) Detailed physics model supports failure diagnosis.}
By pairing task-level verification with simulator-ground-truth
inspection, VABAR helps identify the physical conditions associated
with execution failures. Fig.~\ref{fig:syringe_failure_diagnosis}
illustrates two jammed syringe-disposal episodes, \texttt{t01}
and \texttt{t29}, which terminate with final verification reporting
\texttt{blocked} and 7/8 and 6/8 syringes recorded as posted,
respectively. In both examples, a syringe straddles the lid plane,
with its needle end inside the cavity and its opposite end above
the opening, consistent with incomplete passage into the container.
The surrounding syringes occupy different configurations,
highlighting execution-induced variation: earlier placements
alter the geometry encountered by later insertions.

\noindent\textbf{(d) Skill reuse is extensive across diverse task classes.}
Despite spanning rigid-object manipulation, contact-rich assembly, precision insertion, and deformable-object manipulation, VABAR policies exhibit substantial reuse of underlying skills and tools. Core robot and perception capabilities, including end-effector pose retrieval and SAM3~\cite{carion2025sam}-based segmentation, are shared across all six task classes. Higher-level semantic skills are also reused across tasks with different objects and applications but similar manipulation structure. For example, Tool Hanging and Syringe Packing both rely on skills for held-object motion, functional-feature perception, object registration, and feature mating. This reuse suggests that increasing task diversity does not necessarily require an entirely new set of skills. Instead, VABAR exposes a central challenge for agentic robotics on selecting appropriate skills, parameterizing them for the current geometry and task state, and composing them into effective task-specific behaviors. We hope VABAR will be used by others to facilitate and compare research progress worldwide.

\renewcommand*{\bibfont}{\footnotesize}
\printbibliography

\end{document}